# MLR: A Two-stage Conversational Query Rewriting Model with Multi-task Learning

Shuangyong Song, Chao Wang, Qianqian Xie, Xinxing Zu, Huan Chen, Haiqing Chen
Alibaba Group, Hangzhou 311121, China
{shuangyong.ssy; chaowang.wc; huafan.xqq; patrick.zxx; shiwan.ch; haiqing.chenhq}@alibaba-inc.com

## ABSTRACT

Conversational context understanding aims to recognize the real intention of user from the conversation history, which is critical for building the dialogue system. However, the multi-turn conversation understanding in open domain is still quite challenging, which requires the system extracting the important information and resolving the dependencies in contexts among a variety of open topics. In this paper, we propose the conversational query rewriting model - **MLR**, which is a **M**ulti-task model on sequence **L**abeling and query **R**ewriting. MLR reformulates the multi-turn conversational queries into a single turn query, which conveys the true intention of users concisely and alleviates the difficulty of the multi-turn dialogue modeling. In the model, we formulate the query rewriting as a sequence generation problem and introduce word category information via the auxiliary word category label predicting task. To train our model, we construct a new Chinese query rewriting dataset and conduct experiments on it. The experimental results show that our model outperforms compared models, and prove the effectiveness of the word category information in improving the rewriting performance.

## 1 Introduction

As one of the most natural manner of human–machine interaction, the dialogue systems have attracted much attention recent years. Conversational context understanding plays a key role in building dialogue systems, which aims to recognize the real intention of user from the conversation texts across multi-turn. However, it is still quite challenging in open domain, which requires the system extracting important information and resolving the dependencies in contexts among a variety of open topics. The contexts of multi-turn conversation are redundant, among which only key components are closely related to the latent intentions of users. Moreover, there are complex dependencies among conversational contexts, such as co-reference and rephrase, which further makes challenges for the conversational context understanding.

To decouple the complex dependencies among contexts, there is few work [1-3] consider formulate the conversational context understanding as the query reformulation problem, rewriting the multi-turn conversational queries into a understanding friendly single turn query. However, these methods directly apply sequence-to-sequence models in this task, and treat all contexts equally. Thus, their generation performance is limited without identifying the key information during the query generation.

To deal with the aforementioned issues, we propose a novel query rewriting model based on multi-task learning. To capture the key information during generation, we introduce word category information via the auxiliary word category label predicting task. In the model, a conditional random fields (CRF) layer on top of the Bi-LSTM encoder layer is used to predict the category label of each word. The category vector of each word corresponding to its category label will be added to its hidden state to guide the query generation. Finally, the generation and word category predicting loss are optimized jointly to train the model. To supervisedly train our model, we construct a new Chinese query rewriting dataset from the intelligent personal assistant *AliMe* and conduct experiments on it. The experimental results show the effectiveness of our model and the word category information in improving the rewriting performance.

## 2 Our Model - MLR

### 2.1 Problem Formalization

Given historical queries $Q = \{q_1, q_2, …, q_{n-1}\}$, where $q_i$ is the $i$th ($1 \leq i \leq n-1$) turn of query and $q_n$ is the last query, our model aims to learn a map function $p(q_n, c|Q)$, which can rewrite multi-turn queries $Q$ into a new single turn query $q_n$ and predict the category label $c_i$ of each token $i$.

### 2.2 Model Overview

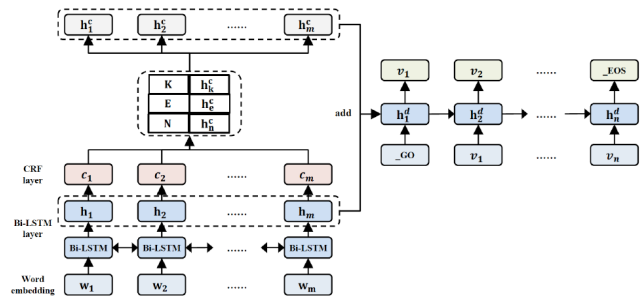

**Figure 1: Model architecture.**

In this subsection, we illustrate our model in detail, whose architecture is depicted in Figure 1.

### 2.2.1 Word representation layer

Assuming $(Q,q_n) = \{w_1, ..., w_m\}$, each token $w_i$ in queries is converted into a continuous representation $e_i$ with a 300-dimensional pre-trained word embedding $e$ in this layer. The word embeddings can be fine tuned during the training process.

### 2.2.2 Bi-LSTM layer

We use two layer Bi-LSTM to encode all queries. Given the input word embedding $e_i$ in current $i$ time, we have the output of the forward and backward LSTM hidden layer unit:

$$h_i^f = \overrightarrow{LSTM}(h_{i-1}^f, e_i) \qquad h_i^b = \overleftarrow{LSTM}(h_{i+1}^b, e_i) \qquad (1)$$

We concatenate the two directional hidden states as the final representation of token $w_i$: $h_i = [h_i^f, h_i^b]$.

### 2.2.3 CRF layer

We introduce a linear chain CRF layer to predict the category label of each token. We define three word category labels: key word label $K$, separation word label $E$, and normal word label $N$, and their corresponding label vectors $\{h_k^c, h_e^c, h_n^c\}$. Given the outputs of Bi-LSTM $h = \{h_1, ..., h_m\}$, the conditional probability of word category label sequence $c$ is:

$$p(c|h;W,b) = \frac{\prod_{i=1}^m \psi_i(c_{i-1}, c_i, h_i)}{\sum_{\tilde{c}_i \in C(h)} \prod_{i=1}^m \psi_i(\tilde{c}_{i-1}, \tilde{c}_i, h_i)} \qquad (2)$$

where $C(h)$ is the set of all possible category labels, $\psi_i(c, \tilde{c}, h_i) = exp(W_{c,c}^T h_i + b_{c,\tilde{c}})$ is the feature function. After predicting the category label $c_i$ of token $w_i$, we add its hidden representation to its category vector $h_i^c$ to form its final representation: $h_i^e = h_i + h_i^c$.

### 2.2.4 Decoder layer

An attention based LSTM decoder is used to predict tokens of the response. At each time step t, given the word embeddings $e_{t-1}$ of previous decoded words and $t-1$ hidden state $h_{t-1}^d$, the decoding process can be depicted as:

$$h_t^d = LSTM(h_{t-1}^d, e_{t-1}) \quad c_t = \sum_{i=1}^m \alpha_{ti} h_i^e$$

$$\alpha_{ti} = softmax(u_{ti}) \quad u_{ti} = V^T \tanh(W_2 \tanh(W_1[h_t^d, h_i^e]))$$

$$o_t = \tanh(W_c[h_t^d, c_t]) \quad y_t = softmax(W_o o_t) \qquad (3)$$

where $y_t$ is the predicted token, and $W_1$, $W_2$, $W_c$, $W_o$ are the weighting parameters.

### 2.2.5 Joint training

We jointly optimize the generation loss $Lg$ and word category predicting loss $Lc$ to train our model:

$$L = Lg + Lc = -\sum_i \log p(y_i) - \sum_j \log p(c_j) \qquad (4)$$

where $p(y_i)$ is the predicted probability of the correct target token $y_i$, $p(c_j)$ is the predicted probability of the correct target label sequence $c_j$.

## 3 Experiments

**Dataset collection**: We collect a Chinese query rewriting dataset from the intelligent assistant *AliMe*, which contains 5,342,981 training data and 50,000 test data. We extract conversations during the same session as a data sample, in which users rewrite the previous queries to get required answers from the chatbot. Tokens of former queries appeared in the last query are labeled as key words. For token "<SEP>" in queries, we label it as separation word. Remaining tokens are labeled as normal words.

**Implementation**: Our model is implemented in Python and *TensorFlow*. In experiments, we use the 300-dimension pre-trained word embeddings based on *fastText* model [4]. We set hidden size to 128 both in encoder and decoder, and learning rate to 0.001. All weights are initialized with the glorot uniform.

**Experimental results**: We compare the following models: the sequence to sequence model (seq2seq), seq2seq with copy net (seq2seq_copy), seq2seq with CRF predicting layer in which the generation and predicting loss are optimized separately (seq2seq_crfs), seq2seq with CRF predicting layer in which the generation and predicting loss are optimized jointly (seq2seq_crfj). We randomly select 500 samples from test data, calculate the BLEU score, the exact match (EM) score [2] and evaluate the rewriting accuracy (ACC) via human evaluation. Table 1 presents the query rewriting accuracy, BLEU and exact match score of different models. It shows that our joint optimizing model seq2seq_crfj outperforms other models. Moreover, it yields a significant improvement over seq2seq, which proves the effectiveness of word category information in improving the rewriting performance.

**Table 1: The query rewriting accuracy, BLEU and exact match score (%) of different models.**

| model | ACC | BLEU-1 | BLEU-2 | BLEU-3 | EM |
|---|---|---|---|---|---|
| seq2seq | 63.87 | 34.96 | 26.90 | 22.06 | 52.93 |
| seq2seq_copy | 67.86 | 40.61 | 28.45 | 22.03 | 54.91 |
| seq2seq_crfs | 68.06 | 41.75 | 31.05 | 23.85 | 56.03 |
| seq2seq_crfj | **84.03** | **45.92** | **33.26** | **26.53** | **58.01** |

## 4 Conclusion

In this paper, we propose a novel conversational query rewriting model MLR, based on multi-task learning, and construct a new large-scale Chinese query rewriting dataset, which will push forward the research in this area after release. In the future, we will try to realize a 'reduced model' to better meet the Queries-per-second (QPS) needs of real online applications.

## REFERENCES
[1] Rastogi, P.; Gupta, A.; Chen, T.; and Lambert, M. Scaling multi-domain dialogue state tracking via query reformulation. NAACL 2019, 97–105.
[2] Ren, G.; Ni, X.; Malik, M.; and Ke, Q. Conversational query understanding using sequence to sequence modeling. WWW 2018, 1715–1724.
[3] Su, H.; Shen, X.; Zhang, R.; Sun, F.; Hu, P.; Niu, C.; and Zhou, J. Improving multi-turn dialogue modelling with utterance rewriter. ACL 2019, 22-31.
[4] Joulin, A., Grave, E., Bojanowski, P., Douze, M., J´egou, H´., and Mikolov, T. Fasttext.zip: Compressing text classification models. arXiv:1612.03651, 2016.